\documentclass{article}

\usepackage{PRIMEarxiv}

\usepackage[utf8]{inputenc} 
\usepackage[T1]{fontenc}    
\usepackage{hyperref}       
\usepackage{url}            
\usepackage{booktabs}       
\usepackage{amsfonts}       
\usepackage{nicefrac}       
\usepackage{microtype}      
\usepackage{lipsum}
\usepackage{fancyhdr}       
\usepackage{graphicx}       
\graphicspath{{media/}}     

\usepackage{float}

\restylefloat{figure}
\restylefloat{table}
\usepackage{algorithm}
\usepackage{algorithmicx}
\usepackage{algpseudocode}
\usepackage{multirow}

\usepackage{mathtools,amssymb}
\title{AGSOA:Graph Neural Network Targeted Attack Based on Average Gradient and Structure Optimization

}

\author{
  Yang Chen, Bin Zhou \\
  School of Computer Science \\
  Qinghai Normal University \\
  Xining, Qinghai, China\\
  \texttt{yangc2753, zhoudat}@163.com} 

\begin{document}
\maketitle

\begin{abstract}
Graph Neural Networks(GNNs) are vulnerable to adversarial attack that cause performance degradation by adding small perturbations to the graph. Gradient-based attacks are one of the most commonly used methods and have achieved good performance in many attack scenarios. However, current gradient attacks face the problems of easy to fall into local optima and poor attack invisibility. Specifically, most gradient attacks use greedy strategies to generate perturbations, which tend to fall into local optima leading to underperformance of the attack. In addition, many attacks only consider the effectiveness of the attack and ignore the invisibility of the attack, making the attacks easily exposed leading to failure. To address the above problems, this paper proposes an attack on GNNs, called AGSOA, which consists of an average gradient calculation and a structre optimization module. In the average gradient calculation module, we compute the average of the gradient information over all moments to guide the attack to generate perturbed edges, which stabilizes the direction of the attack update and gets rid of undesirable local maxima. In the structure optimization module, we calculate the similarity and homogeneity of the target node's with other nodes to adjust the graph structure so as to improve the invisibility and transferability of the attack. Extensive experiments on three commonly used datasets show that AGSOA improves the misclassification rate by 2$\%$-8$\%$ compared to other state-of-the-art models.
\end{abstract}


\section{Introduction}\label{Section:Intro}

With the application of deep learning on graph data, Graph Neural Networks(GNNs) have shown remarkable performance \cite{A1}. GNNs have been applied to node classification \cite{ WOS:001214105000046,WOS:001170500200001}, graph classification \cite{WOS:001175221000046,WOS:001124222100012} and link prediction \cite{A4,WOS:001155057000016} tasks by aggregating nodes' neighborhood information to learn structre and feature information of graph data. Recent studies have shown that GNNs inherit the vulnerability of deep learning, where the adversary adds small perturbations (structures or features) to the graph leading to incorrect predictions, which in turn have unpredictable consequences \cite{Transferable,Nettack}. For example, in social networks, an attacker can add some non-existent users and establish fake social relationships with key users, which can lead to confusion on information dissemination, damage the trust relationship between users, and affect the decision-making of key users.

Graph adversarial attacks can be categorized as Targeted Attacks and Untargeted Attacks based on the attack objectives \cite{liu2022towards,dai2022targeted}. In targeted attacks, the attacker usually chooses one or several nodes as the target nodes. GNNs misclassify the target nodes by modifying the links between the target nodes and other nodes \cite{FGA}. For example, Wang et al. \cite{wang2023revisiting} found that the gradient estimates are often noisy, which leads to a ineffective attack, and proposed a targeted attack that mentions masking the attack noise by modifying the structure to make the target nodes misclassified. Untargeted attacks aim to cause global nodes misclassification by interfering with the entire graph structure \cite{chen2022practical}. Liu et al. \cite{liu2022towards} demonstrated theoretically that negative cross-entropy tends to generate more significant gradients from nodes with lower confidence in the labeled categories, and proposed an untargeted graph structure attack via Gradient Debias, aiming to cause global nodes to fail by modifying the structure.

Targeted attacks have the advantage of invisibility and effectiveness compared to untargeted attacks \cite{lin2023exploratory,chen2022graphfool,NAG}. \textbf{Invisibility}: the target attacks are performed on specific nodes or edges, and does not require changes to the structure of the entire graph, so the attack is more invisible and not easy to be detected and defended. \textbf{Effectiveness}: Attackers can select target nodes or edges to achieve more precise attack targets, which is more flexible and effective compared with traditional global attacks.
Attackers tend to attack the core nodes in the network, so targeted attacks are often applied in practical scenarios \cite{NAG}. In addition, in recommender systems, targeted attacks can also have a positive impact on the system \cite{nguyen2023poisoning}.
For example, targeted attacks can be used to accurately direct targeted users to watch correctly oriented content or views in advertisement recommendations and public recommendations.

In recent years, many graph adversarial attack methods have been proposed \cite{xing2023clean,chen2023empirical}. Since GNNs training is achieved by continuously optimizing the gradient, the gradient provides an understanding of the behavior of the GNNs. The attacker can use the gradient information to understand how the GNNs classify or predict the input data   \cite{WOS:001139144400045,2022arXiv220812815L,AFGSM}. Therefore, most attacks are based on gradients  \cite{xu2024attacks,tao2024black,hu2023hyperattack}.

\begin{figure}[h]
	\centering
	\includegraphics[width=1\textwidth]{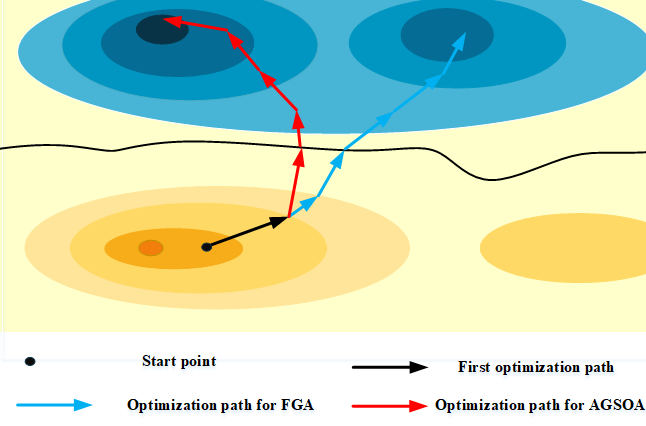}
	\caption{Illustration of two gradient attack optimization paths. FGA attacks along the direction of the gradient of the previous iteration. AGSOA accumulates the average of the gradient of all previous moments of the iteration to attack. The more gradients that are accumulated, the easier it is for the attack to find a global optimum.}\label{fig1}
\end{figure}

Although existing gradient attacks address some of the difficulties of graph adversarial learning, there are still problems that remain unsolved: \textbf{Easily fall into the local optimum}. Many gradient attacks use the idea of greedy way and iteratively attack the edge or feature corresponding to the absolute maximum value of the current gradient, which is easy to make the attack fall into the local optimum resulting in the attack obtaining a locally optimal solution \cite{MGA}. Fig. \ref{fig1}  illustrates the FGA optimization path, FGA is one of the first works to propose the use of GCN gradient information to generate adversarial samples \cite{FGA}.
FGA is effective in degrading the performance of GNNs but fails to find the attack's global optimum. \textbf{Poor attack invisibility}. For targeted attacks, many attacks only consider the effectiveness of the attack, i.e., it costs more to misclassify the nodes, and do not notice the damage caused by the attack to the original graph, such as node similarity and homogeneity, which makes the attack easy to be detected by the defense model.

To address the above difficulties, in this work, we propose a target attack on GNNs based on average gradient and structure optimization, called AGSOA. AGSOA consists of two components, average gradient computation and structre optimization.  In the average gradient section, we use the average gradient method to accumulate the average gradient direction for all moments, which guides the attack to update towards the attack's global optimum and avoids the attack from falling into a local optimum. Fig. \ref{fig1}  illustrates the AGSOA optimization path, where AGSOA improves the performance of the attack by using the average gradient to find the perturbation of the current iteration. We use structre optimization modules to ensure the imperceptibility of attacks. Specifically, AGSOA computes the similarity and homogeneity of the target nodes with other nodes. Then AGSOA rewires them using the TOP-K overlapping technique. Extensive experiments have shown that AGSOA achieves better misclassification rates and transferability.

The main contributions of this paper are as follows:

$\bullet$  We propose a novel attack AGSOA based on average gradient and structure optimization, which is oriented towards the target attack of GNNs.

$\bullet$ We improve the performance of the attack by accumulating the average value of the gradient to avoid the attack falling into a local optimum.

$\bullet$ We optimize the graph structure through node similarity and node homogeneity metrics to improve attack performance while ensuring attack invisibility.

$\bullet$ We compare AGSOA with state-of-the-art target attack methods in three types of common datasets, and the results show that AGSOA achieves the significant improvement.

The rest of the paper is organized as follows: Section \ref{S2}  briefly reviews the works on GNNs adversarial attacks. Section \ref{S3} provides mathematical expressions for GNNs training, attack aims and gradient modification rules. In Section \ref{S4}, the components of AGSOA are described in detail. In Section \ref{S5}, we describe the experimental setup, show and analyze the experimental results. In Section \ref{Sec:conclusion}, conclusions are reported.

\section{Related Work}\label{S2}

The works on graph adversarial attacks can be classified into optimization strategy-based attacks, reinforcement learning-based attacks and gradient-based attacks in terms of graph data perturbation. In this section, we review the classical GNNs adversarial attacks from the above classifications.

(1) Optimization Strategy-based attacks. The attacker modifies the graph data by using an optimization strategy to make the GNNs produce false predictions \cite{liu2021neighbor,tao2021single}. Liu et al. \cite{liu2024revisiting} designed an edge prioritization detector to generate a new prioritization metric to enhance the edge perturbation effect. Zhang et al. \cite{zhang2024maximizing} proposed a global node injection attack framework that uses a susceptible reverse influence sampling strategy and contrast loss to optimize the attack target by updating edges and nodes information. Chen et al. \cite{chen2023feature} found that feature triggers destroy the feature space of the original dataset and proposed an adaptive method to improve the performance of the backdoor model by adjusting the graph structure. The literature \cite{chen2024imperceptible} proposed an imperceptible graph injection attack that uses the homogeneous imperceptibility constraint to improve the camouflage of fake nodes. Fang et al. \cite{GNAI} proposed a fake node injection attack that uses feature statistics and evolutionary perturbation information obtained from a genetic algorithm to generate features and select neighbors for fake nodes. Sheng et al. \cite{sheng2021backdoor} proposed a backdoor attack on GNNs based on subgraph triggers, which designs triggers based on the characteristics of the sample data and uses a random graph generation algorithm to obtain subgraph triggers.

(2) Reinforcement Learning-based attacks. Attackers use reinforcement learning algorithms to optimize their attack strategies and interfere with the learning of the GNNs \cite{sun2022adversarial}. RL-s2v uses reinforcement learning to determine the best strategy for perturbing graph structure and features \cite{dai2018adversarial}. RL-s2v learns strategies by observing the current state of the graph, the actions taken, and the changes in predicted outcomes. The literature \cite{yang2023gaa} proposed a graph adversarial attack that uses reinforcement learning to sequentially generate features and links for fake nodes without modifying existing nodes or edges. Tao et al. \cite{tao2024black} proposed a black-box attack based on hierarchical reinforcement learning, which divides the dynamic graph structure perturbation into three subtasks and transforms them into a continuous decision-making process to achieve the attack. Ju et al. \cite{ju2023let} modeled the node injection attack as a Markov decision process and proposed a reinforcement learning attack based on advantage actor critic. Sharma \cite{sharma2023task} devised an effective heuristic attack to cause node neighborhood distortion by combining graph homomorphic networks with deep Q-learning, which effectively degrades prediction performance significantly. 

(3) Gradient-based attacks. The attacker uses the gradient information to modify the features or edges of the nodes to reduce the accuracy of the GNNs \cite{jain2024stronger,li2022revisiting}. NETTACK is the first work in the field of graph adversarial learning, which utilizes gradient information to undermine the structre integrity of graphs, thus degrading the performance of GNNs \cite{Nettack}. Zhang et al. \cite{zhang2023minimum} proposed a minimum budget topology attack that adaptively finds the minimum perturbation to successfully attack each node. Zhao et al. \cite{zhao2024hgattack} used the gradient information of the surrogate model to generate perturbation edges to reduce the performance of heterogeneous graph neural networks.The literature \cite{chen2022graphfool} proposed a target labeling adversarial attack against graph embeddings, which generates perturbation graphs by categorizing boundary and gradient information. Lin et al. \cite{lin2023exploratory} proposed a gradient-based graph adversarial attack, which avoids the problem of traditional gradient attacks falling into local optima through three modules: generation, evaluation and reorganization. Shang et al. \cite{shang2023transferable} used a gradient-guided attack for edge perturbation to explore the robustness of heterogeneous graph neural networks.

Gradient attacks use the gradient information of GNNs, the attacker can more easily understand the model's sensitivity to the input data, and thus better guide the process of generating adversarial samples.
In contrast, graph optimization strategy attacks and graph reinforcement attacks require more black-box operations, and it is less easy for attackers to understand the specific behavior of the model. In addition, graph optimization strategy attacks may require more computational resources and iterative steps, and graph reinforcement attacks may require a more complex strategy search process, which is not easy to implement in practical scenarios. In this paper, we propose a gradient attack for GNNs and verify the effectiveness of the model in experiments.

\section{Preliminaries}\label{S3}
This section gives definitions related to this paper, including the definition of graphs, GNNs definition. In addition, gradient attack rules are also introduced. The common notations used in this paper are given in Table \ref{notations}.

\begin{table}[!ht]
	
	\begin{center}
		\caption{Notations frequently used in this paper and their corresponding descriptions.}\label{notations}%
		\resizebox{\linewidth}{!}{
			\begin{tabular}{cc|cc}
				\toprule
				Notation & Description & Notation & Description\\
				\midrule
				$G$ & Clean graph &   $G^{'}$ & Perturbation graph\\
				$A$ & Adjacency matrix &   $A^{'}$ & Perturbation adjacency matrix\\
				$V$ & Set of nodes &   $n$ & Number of nodes\\
				$E$ & Set of edges &   $m$ & Number of edges\\
				$X$ & Set of node features &   $d$ & Dimension of the node features\\
				${f_\theta }$ & Graph neural network model &   $\Delta$ & Attack budget\\
				$y$ & True label  of node &  ${{\widehat y}}$	 & Predicted label  of node\\
				${f_\theta }$ & Graph neural network model &    $\mu $  & Momentum factor\\
				$ {B}$ & Gradient matrix &   ${\bar B}$ & Average gradient matrix\\
				$ Fs$ & Node feature similarity &   $Ho$ & Node homogeneity\\
				\midrule
		\end{tabular}}
	\end{center}
\end{table}

\subsection{Graph Definition}\label{3.1}

Given a attribute graph $G = (V,E,X)$, Where $V = \{ {v_1},{v_2},...,{v_n}\} $ is the set of nodes, $n$ is the number of nodes. $E = \{ {e_1},{e_2},...,{e_m}\} $ is the set of edges, $m$ is the number of edges.  We use the adjacency matrix $A \in {\{ 0,1\} ^{n \times n}}$ to represent the adjacency of the nodes in the graph $G$.
There is a connecting edge between node $i$ and node $j$ when ${A_{i,j}} = 1$, otherwise there is no connection.
$X \in {\mathbb{R}^{n \times d}}$ is the set of node features and $d$ represents the dimension of the node features.

\subsection{Graph Convolutional Neural Networks Definition}\label{3.2}

GNNs are used in many tasks, this paper focuses on the node classification task and the main model used is Graph Convolutional Network (GCN) \cite{GCN}. An attribute graph $G$ is input into the GCN, which performs node-level prediction or learning by learning the relationships between nodes. Usually, GCN will contain many graph convolutional layers, which can realize multi-level information propagation and feature learning. The output of each layer can be used as the input of the next layer, enabling the model to gradually learn more complex features of graph.

At the $l$-th iteration, the output of the GCN can be expressed as:

\begin{equation}\label{eq1}
{H^l} =  ReLU (\widetilde A{H^{l - 1}}{W^l}).
\end{equation}

Where $ReLU$ is a common nonlinear activation function for GNNs. $\widetilde A = {\widehat D^{ - \frac{1}{2}}}\widehat A{\widehat D^{ - \frac{1}{2}}}$ is the normalized adjacency matrix, $\widehat A = A + I$ is a self-looping adjacency matrix, $\widehat D$ is the node degree diagonal matrix of $\widehat A$. ${H^l}$  and ${H^{l - 1}}$ represent the output vectors of the GCN at  $l$-th and $l-1$-th layers, respectively. When $l = 0$, ${H^0} = X$.  ${W^l}$ is the set of learnable parameters of the GCN at  $l$-th layer.

The output of a GCN with $K$ layers can be represented as:

\begin{equation}\label{eq2}
Z = {f_\theta }(A,X) = softmax (\widetilde A...{\mathop{\rm Re}\nolimits} LU(\widetilde AX{W^1})...{W^k}).
\end{equation}

Usually, the training is completed by updating the parameters $ \theta $ with the cross-entropy loss function ${L_{tra}}(\theta ;X,A)$. 

\begin{equation}\label{eq3}
{L_{tra}}(\theta ;X,A) =  - \sum\limits_{i \in {V_{train}}} {{y_i}\ln {{\widehat y}_i}} ,{\rm{  }} \quad s.t.  \quad {\rm{ }}{\widehat y_i} = \arg \max ({Z_{i,:}}){\rm{ }}.
\end{equation}

Where ${V_{train}}$ is the set of known labeled nodes,  ${y_i}$ and ${{\widehat y}_i}$ are the true and predicted labels of node $i$, respectively.

\subsection{Threat Model}\label{3.3}
AGSOA sets the targeted attacks which aims to modify the edges of the target nodes resulting in misclassification.  In other words, the goal of the attack is to reduce the classification accuracy in the set of target nodes. The attack function can be set as:

\begin{equation}
\max _{G^{\prime}} \sum_{i \in V_{\text {Tar }}}\left\{\hat{y}_i \neq y_i\right\} \text {, s.t. }\left\|A^{\prime}-A\right\|_0 \leq \Delta.
\end{equation}

where $G^{\prime}$ is the perturbed graph after the attack, and ${A^{\prime}}$ is the adjacency matrix that joins the perturbed edges.  ${V_{Tar}}$ is the set of target nodes.  $\Delta $ is the attack budget to ensure the invisibility of the attack, and if an undirected graph is studied, the budget is set to $2\Delta $ .

\subsection{Gradient Modification Rule}\label{3.4}
AGSOA uses gradient information to modify edges, aiming to increase the training loss of GNNs.  Larger absolute values of the gradient cause more damage to the GNNs, so we use the gradient modification rule to modify the edge. The core idea of the gradient modification rule is to add and remove edges based on the gradient, adding links with the largest positive gradient and deleting links with the smallest negative gradient.
The gradient modification rule can be formalized as follows:

\begin{equation}\label{eq5}
\left\{\begin{array}{l}
\text { Add } e_{(i, j)}, \text { s.t. } B_{i, j}^{(k)}>0 \text { and } A_{i, j}^{\prime (k)}=0. \\
\text { Delete } e_{(i, j)}, \text { s.t. } B_{i, j}^{(k)}<0 \text { and } A_{i, j}^{\prime (k)}=1.
\end{array}\right.
\end{equation}

Where ${B^k} \in {\mathbb{R}^{n \times n}}$ is the gradient matrix.

section{Preliminaries}\label{S3}
This section gives definitions related to this paper, including the definition of graphs, GNNs definition. In addition, gradient attack rules are also introduced. The common notations used in this paper are given in Table \ref{notations}.

\begin{table}[!ht]
	
	\begin{center}
		\caption{Notations frequently used in this paper and their corresponding descriptions.}\label{notations}%
		\resizebox{\linewidth}{!}{
			\begin{tabular}{cc|cc}
				\toprule
				Notation & Description & Notation & Description\\
				\midrule
				$G$ & Clean graph &   $G^{'}$ & Perturbation graph\\
				$A$ & Adjacency matrix &   $A^{'}$ & Perturbation adjacency matrix\\
				$V$ & Set of nodes &   $n$ & Number of nodes\\
				$E$ & Set of edges &   $m$ & Number of edges\\
				$X$ & Set of node features &   $d$ & Dimension of the node features\\
				${f_\theta }$ & Graph neural network model &   $\Delta$ & Attack budget\\
				$y$ & True label  of node &  ${{\widehat y}}$	 & Predicted label  of node\\
				${f_\theta }$ & Graph neural network model &    $\mu $  & Momentum factor\\
				$ {B}$ & Gradient matrix &   ${\bar B}$ & Average gradient matrix\\
				$ Fs$ & Node feature similarity &   $Ho$ & Node homogeneity\\
				\midrule
		\end{tabular}}
	\end{center}
\end{table}

\section{Average Gradient and structre Optimization Attack}\label{S4}

\begin{figure}[hpt]
	\centering
	\includegraphics[width=1\textwidth]{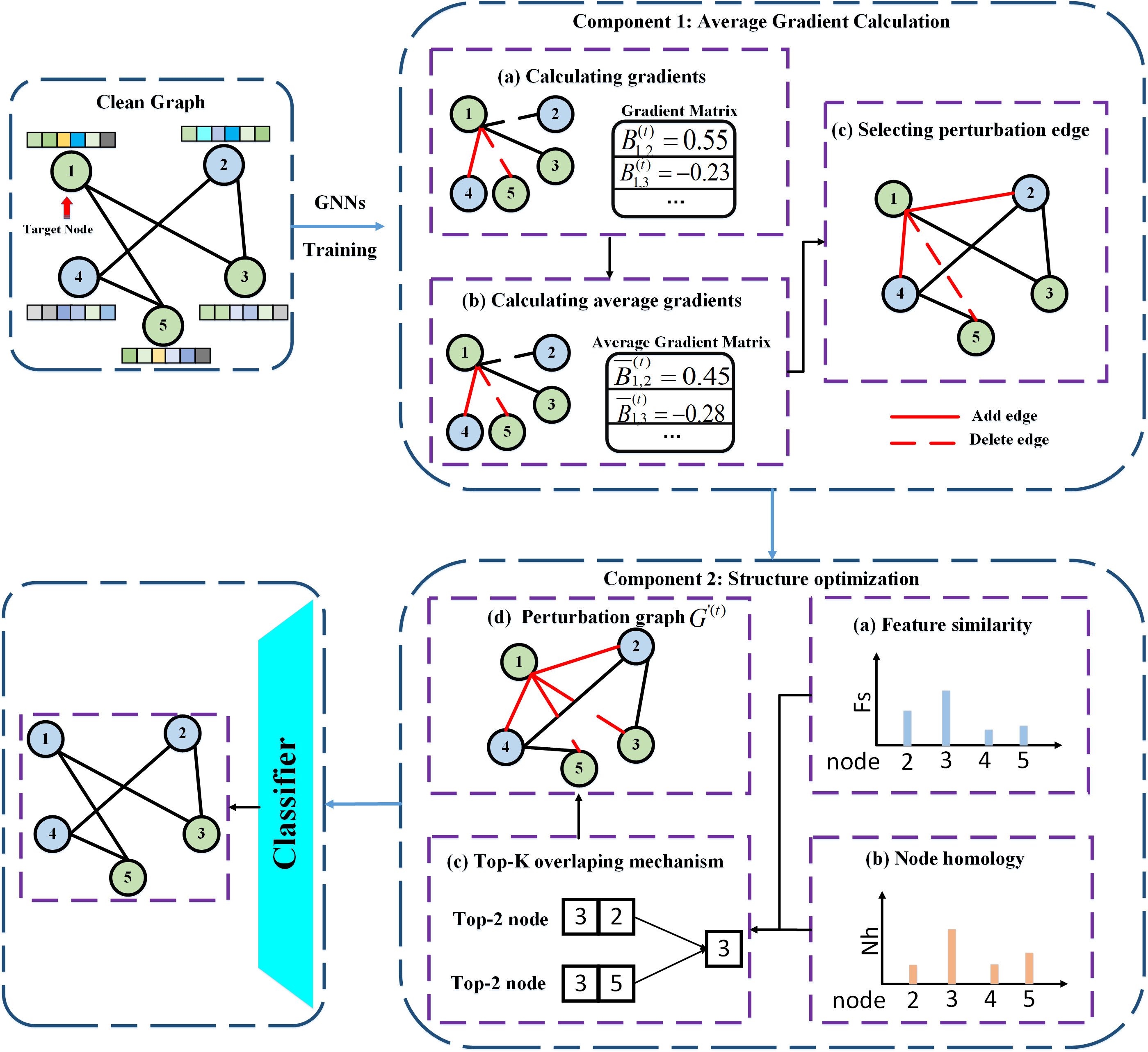}
	\caption{AGSOA Overall Framework. AGSOA consists of two components: average gradient computation and structre optimization. At the $t$-th iteration, the average gradient ${\bar B^{(t)}}$ is obtained by accumulating the gradients of the previous $t$ moments, and the attack uses the gradient modification rule to add or delete edges. In the structure optimization component, we compute the feature similarity and node homogeneity between the target node and other nodes.  AGSOA then use the TOP-K overlapping mechanism to select the perturbation edges to generate the perturbation graph $G^{\prime (t)}$. The final perturbation graph $G^{\prime}$ is classified to get the predicted label of the target node, if the predicted label of the target node is different from the real label means the attack is successful.}\label{fig2}
\end{figure}

In this section, we describe the overall framework of AGSOA in detail. It consists of two components: average gradient computation and structure optimization, as shown in Fig. \ref{fig2}.

\subsection{Average Gradient Calculation}\label{4.1}

Studies on gradient optimization such as MGA \cite{MGA} and NAG \cite{NAG} results show that good gradient attack strategies can make the attack avoid falling into a local optimum thus improving the attack performance.
However, these attacks cannot obtain more robust gradient information by using the gradient information (i.e., momentum) from the previous moment to guide the attack. Literature \cite{liu2023enhancing} results show that accumulating multiple moments of gradient attacks exhibit better performance than using only the gradient of the previous. The above results show that accumulating multiple gradient moments helps the attacker to overcome bad local optima and generate better perturbations \cite{WOS:001133324200016}. Inspired by this, we propose an average gradient attack strategy using the forward-looking nature of momentum gradient.

Specifically, AGSOA is trained using GNNs to obtain the gradient  ${B^t}$ of the loss function with respect to the adjacency matrix $A$, which can be described as:

\begin{equation} \label{eq6}
B^{(t)}=\frac{\partial L_{\text {tra }}\left(\theta ; X, {A}^{\prime(t)}\right)}{\partial {A}^{\prime(t)}}=\nabla_{{A}} L_{\text {tra }}\left(f_\theta\left(\theta ; {X}, {A}^{\prime(t)}\right)\right).
\end{equation}

where $t$ is the number of iterative attacks and ${A}^{\prime(t)}$ is the adjacency matrix generated at the $t$-th attack. When $t = 0$, ${A}^{\prime(0)}$ is the adjacency matrix in the clean graph, i.e., ${A}^{\prime(0)}=A$.

The aim of AGSOA is to accumulate gradients from more moments to improve the transferability of the attack.
Specifically, the previous gradients are added to the current gradient in each iteration and then averaged to get the average gradient at moment $t$. The average gradient ${\bar B^{(t)}}$ at time $t$ can be expressed as follows.

\begin{equation}\label{eq7}
{\bar B^{(t)}} = \left\{ {\begin{array}{*{20}{c}}
	{{B^{(0)}},t = 0.}\\
	{\frac{1}{{t + 1}}[{\nabla _{{A^{'(t)}}}}L_{\text {tra }}({f_\theta }(\theta ;X,{A^{'(t)}})) + \sum\limits_{i = 0}^{t - 1} {{\nabla _{{A^{'(i)}}}}} L_{\text {tra }}({f_\theta }(\theta ;X,{A^{'(i)}}))],t \ge 1.}
	\end{array}} \right.
\end{equation}

When $t = 0$, ${\overline B ^{(0)}} = {B^{(0)}} = {\{ 0\} ^{n \times n}}$.

As the number of iterations $t$ increases, AGSOA uses  more and more moments gradient information to guide the attack to generate perturbed edges. The $t$-th attack adjacency matrix can be formalized as follows:

\begin{equation}\label{eq8}
A_{nor}^{^{'(t)}} = {A^{'(t - 1)}} + \mu {\overline B ^{(t)}}.
\end{equation}

where $\mu $ is the momentum factor, and when  $\mu $  is large, the generated perturbations mainly depend on the gradient information.
$A_{nor}^{^{'(t)}}$ is a normalized adjacency matrix with continuous values.

The perturbation edges generated at the moment $t+1$ can be obtained through ${\bar B^{(t)}}$. The calculation procedure is as follows:

\begin{equation}\label{eq9}
{\bar B^{(t + 1)}} = \mu {\bar B^{(t)}} + \frac{{{{\bar B}^{(t)}}}}{{||{{\bar B}^{(t)}}|{|_1}}}.
\end{equation}

where $|| \cdot |{|_1}$ denotes the L1 norm.

In the gradient attack, the edge with the larger absolute value of gradient can have a larger impact on the optimization direction of the GNNs objective function. Therefore, AGSOA attacks the edge with the largest absolute value of gradient in each iteration.

\begin{equation}\label{eq10}
\bar B_{i,j}^{^{(t + 1)}} = \mathop {\arg \max }\limits_{{v_i},{v_j} \in V} (|{\bar B^{^{(t + 1)}}}|).
\end{equation}

\begin{equation}\label{eq11}
A_{i,j}^{'(t + 1)} = A_{i,j}^{'(t)} + I(\bar B_{i,j}^{^{(t + 1)}}).
\end{equation}

where $I(x)$ is the gradient sign, if nodes $i$ and $j$ satisfy the gradient modification rule Eq. \ref{eq5}, $I(x) = 1$ when $x > 0$, otherwise $I(x) = -1$.

The above process is repeated continuously until the average gradient module ends at $t=T$, $T$ is the number of gradient modifications.

\subsection{Structure Optimization}\label{4.2}

AGSOA has no perturbation limits in the average gradient module, we  use structure optimization strategies to ensure attack stealth. Previous studies have shown that two nodes with links tend to have similar features, i.e., two nodes with embedding approximation in the network \cite{chen2023feature}. Therefore, AGSOA uses node similarity and node homogeneity metrics to modify the graph structure to make the target nodes dissimilar to neighboring nodes.

Both node similarity and node homogeneity metrics are can describe the similarity between nodes. Specifically, node $i$ similarity is defined as:

\begin{equation}\label{eq12}
F{s_{[i]}} = {X_i} - {X_{tar}}.
\end{equation}

where ${X_{tar}}$ is the target node feature. The lower the absolute value of $F{s_{[i]}}$, it means that node $i$ is more similar to the target node.

Node $i$ homogeneity is defined as follows.

\begin{equation}\label{eq13}
H{o_{[i]}} = sim({r_i},{X_{tar}}), \quad {r_i} = \frac{{{X_i}}}{{\sqrt {{d_{tar}}} \sqrt {{d_i}} }}.
\end{equation}

Where, $d_{tar}$ is the target node degree, the larger $Ho$ indicates the greater homogeneity between nodes.

Due to the difference in the range of values taken by $Fs$ and $Ho$, we use the Top-K overlaping mechanism to find the most similar nodes, as shown in Fig.\ref{fig2} (c). Specifically, we select the top $K$ nodes with the smallest and largest absolute values in $|Fs|$ and $Ho$, respectively. AGSOA then select the similar nodes that occur simultaneously in $|Fs|^{k}$ and $Ho^{k}$. If there is no connecting edge between the target node and the similar node then add edge operation is performed and conversely delete edge operation is performed.

To ensure the invisibility of the attack, we set the change of the total degree of the nodes before and after the attack to remain within a certain range.

\begin{equation}
\mid d_G-d_{G^{\prime}}\mid\leq\Delta, \quad \Delta=\alpha d_{tar}.
\end{equation}

where ${d_G}$ and $d_{{G^{\prime}}}$ are the total node degrees of the clean and perturbed graphs, respectively. $\alpha$ is the budget hyperparameter, which is used to control the attack budget.

\subsection{Algorithm and Time Complexity}\label{4.3}
The pseudo-code of AGSOA is given in Algorithm \ref{alg1}.

\begin{algorithm}[ht!]  
	\renewcommand{\algorithmicrequire}{\textbf{Input:}}
	\renewcommand{\algorithmicensure}{\textbf{Output:}}
	\caption{AGSOA}  
	\label{alg1}
	\begin{algorithmic}[1] 
		\Require Graph dataset $G = (V,E,X)$, Set of target nodes $V_{Tar}$, Constraint factor $\varepsilon $, Budget $ \Delta $,  Momentum factor $\mu$, Number of gradient modifications $T$
		\Ensure  Perturbation graph dataset $G = (V,E^{\prime},X)$, Perturbation of adjacency matrix $A^{'}$
		
		\For{$v \in {V_{Tar}}$} 
		
		\State \textbf{Initialization:} $t=0$, $B ^{(0)}$, $A ^{(0)}=A$
		\While {$t<T$ } \Comment{Module 1: Average Gradient Calculation}
		\State Obtain the normalization matrix through $A_{nor}^{^{'(t)}}$ Eq. \ref{eq8}
		\State Calculate the gradient matrix through $B^{(t)}$ Eq. \ref{eq6}
		\State Calculate the average gradient matrix through $\bar B^{(t)}$ Eq. \ref{eq7}
		\State Update the average gradient matrix ${\bar B^{(t + 1)}}$ through Eq. \ref{eq9}
		\State Generate a perturbed adjacency matrix $A_{i,j}^{'(t + 1)}$ through Eq. \ref{eq5}, \ref{eq10} and \ref{eq11}
		\EndWhile

		\While{$\mid d_G-d_{G^{\prime}}\mid>\Delta$ } \Comment{Module 2: Structure Optimization}
		\State Calculate node feature similarity $Fs$  through Eq. \ref{eq12}
		\State Calculate node homogeneity $Ho$  through Eq. \ref{eq13}
		\State Generate a perturbed adjacency matrix $A_{i,j}^{'}$ by  the Top-K overlaping mechanism
		\EndWhile
		\EndFor

	\end{algorithmic}
\end{algorithm}

\textbf{Complexity Analysis.} We analyze the time complexity of AGSOA, pre-training the model with GCN as an example. AGSOA contains modules for average gradient computation and structure optimization. (1) In the average gradient computation module, AGSOA needs to be trained and backpropagated using GCN, and the time complexity is $o({n_{tra}}d||X||)$,  where ${n_{tra}}$ is the number of training sessions and $d$ is the feature matrix dimension. The time complexity of computing the average gradient is small and is again ignored here. Thus the average gradient computation module time complexity is $o({n_{tra}}d||X||)$. In the structure optimization module, AGSOA needs to calculate the similarity and homogeneity between the target node and other nodes with a time complexity of $o(n)$, $n$ is the number of nodes, and the rest of the operations do not take much time. Therefore, the time complexity of the structure optimization module is $o(n)$.

In summary, the time complexity of AGSOA is $o({n_{tra}}d||X|| + n)$.

\section{Experiments}\label{S5}

This section describes the datasets, GNNs model, baseline, hyperparameter settings, and metrics. Finally, we show experimental results to validate the effectiveness of AGSOA.

\subsection{Datasets}\label{5.1}

We evaluate AGSOA on three real and commonly used datasets, i.e., Cora \cite{mccallum2000automating}, Cora-ML  \cite{bojchevski2017deep}, and Citeseer \cite{giles1998citeseer}. In the three citation datasets, nodes represent articles and edges represent citation relationships. In addition, all three datasets are attribute graphs, details of which are given in Table \ref{tab1}.

\begin{table}[h]
	\begin{center}
		\caption{ Statistics of three datasets. We use these three datasets to verify the performance of AGSOA in the node classification task.}\label{tab1}%
		\begin{tabular}{ccccc}
			\toprule
			Datasets & $\#$ Nodes & $\#$ Features &$\#$ Edges & $\#$ Classes  \\
			\midrule
			Cora     & 2708    & 1433 & 5429  & 7           \\
			Cora-ML     & 2995    & 2879  & 8416 & 7              \\
			Citeseer & 3312    & 3703  & 4715  & 6             \\
			\midrule
		\end{tabular}
	\end{center}
\end{table}

\subsection{GNNs Model}\label{5.2}

In our experiments, we mainly use GCN to verify the effectiveness of AGSOA. In addition, we also use SGC and ChebNet to verify the transferability. The three GNNs models are described in detail as follows:

\textbf{GCN} \cite{GCN}: GCN is one of the most classical GNNs. The core idea is to use the adjacency matrix of the graph to propagate and aggregate the feature information of the nodes, and to learn the node representations through multi-layer graph convolution operations.

\textbf{SGC} \cite{SGC}: SGC is a simplified version of GCN. SGC simplifies the computation and reduces the parameters by removing nonlinear activation functions and normalization operations.

\textbf{ChebNet} \cite{ChebNet}: ChebNet uses the property of Chebyshev polynomials to approximate graph convolution operations by a small number of polynomial coefficients, reducing the amount of computation while maintaining accuracy.

\subsection{Baselines}\label{5.3}

Our proposed AGSOA is the targeted attack, so the baseline models can all be oriented towards targeted attack as detailed below:

\textbf{Random}: Random is the simplest type of attack, which randomly modifies the edges between the target nodes and other nodes with probability $p$.

\textbf{GradArgmax} \cite{dai2018adversarial}: GradArgmax is a gradient-based adversarial attack method. GradArgmax assigns a learnable weight to each edge, and selects the edge with a large absolute value of the derivative for modification after backward derivation.

\textbf{FGA} \cite{FGA}: FGA extracts the gradient information of edges from GNNs and then selects the node pair with the largest absolute value of gradient to update the perturbation graph.

\textbf{MGA} \cite{MGA}: MGA uses the momentum gradient method to improve the effectiveness of gradient attacks.

\textbf{TUA} \cite{dai2022targeted}: TUA enhances the attack with a small number of fake nodes connected to the target nodes.

\textbf{NAG-R} \cite{NAG}: NAG-R generates perturbed edges using Nesterov accelerated gradient attack and Rewiring optimization methods.

Where Random, FGA, MGA and NAG-R can also be extended to the untargeted attacks.

\subsection{Parameter setting and metric}\label{5.4}

\textbf{Parameters}. All the experiments we conducted were executed on the Pytorch computing framework and run with 2 parallel NVIDIA GeForceGTX1080Ti GPUs. For GNNs, the number of layers is set to 2 for all models, the activation function is set to ReLU, the number of hidden layers is set to 64, and the optimizer uses Adam with a learning rate of 0.001. The momentum factor $\mu $ in AGSOA is 0.6 and the budget hyperparameter is 0.2.

\textbf{Metric}. To evaluate the performance of our model in target attack, the Misclassification Rate(MR) is used as an evaluation metric with the following mathematical expression:

\begin{equation}
MR = \frac{{\sum\limits_{i \in {V_{Tar}}} {\{ {{\widehat y}_i} \ne {y_i}\} } }}{{|{V_{Tar}}|}}.
\end{equation}

\subsection{Experimental Results}\label{5.5}
\subsubsection{AGSOA Attack Performance}\label{5.5.1}

In this section, we validate the effectiveness and transferability of our proposed model under several types of comparative models based on three classes of GNNs.
Table \ref{tab2} demonstrates the MR for several types of state-of-the-art models. 

Table \ref{tab2} shows that AGSOA can achieve the highest MR. For example, in Cora, the MR of AGSOA is 7$\%$ and 11$\%$ higher than the TUA and NAG-R, respectively. The results are the same in Cora-ML and Citeseer datasets. The above results illustrate that our proposed method enables GNNs to easily distinguish between target nodes and other nodes, and can successfully misclassify target nodes.

In addition, we extend AGSOA to SGC and ChebNet to verify the transferability. The results show that our proposed attack achieves the best misclassification accuracy in all types of GNNs.  For example, taking SGC as an example, the MR of FGA, MGA, TUA, NAG-R and AGSOA are 59.8$\%$, 60.7$\%$, 78.9$\%$, 78.6$\%$ and 82.4$\%$ respectively in Cora. The higher the MR, the better the attack performance.  In ChebNet, the results are the same. Therefore, AGSOA can achieve high MR in all types of GNNs, i.e., AGSOA has good transferability.

\begin{table}[!ht]
	\caption{MR ($\%$) for several types of targeted attacks, where higher MR represent better attack performance, with the best results bolded. The results are the average of 10 runs.}\label{tab2}%
	\begin{center}
		\resizebox{\linewidth}{!}{
			\begin{tabular}{c|ccc|ccc|ccc}
				\toprule
				Model&\multicolumn{3}{c}{GCN}&\multicolumn{3}{c}{SGC}&\multicolumn{3}{c}{ChebNet}\\
				\midrule
				Datasets&Cora&	Cora-ML&	Citeseer&Cora&	Cora-ML&	Citeseer&Cora&	Cora-ML&	Citeseer\\
				\midrule
				Random&43.2&47.4&37.4&42.2&46.5&38.7&42.0&46.5&	37.5\\
				GradArgmax&59.6&64.2&60.2&58.1&63.2&65.1&58.5&63.2&	64.2\\
				FGA&58.1&66.6&67.4&59.8&67.4&66.4&57.2&65.2&65.8\\
				MGA&63.8&67.6&68.5&60.7&68.1&67.5&57.9&66.3&67.1\\
				TUA&78.4&79.3&70.2&76.1&78.9&71.2&75.1&78.2&71.8\\
				NAG-R&76.5&75.2&71.8&73.4&78.6&71.2&73.5&78.6&70.2\\
				AGSOA&\textbf{81.3}&\textbf{83.6}&\textbf{73.5}&\textbf{81.8}&\textbf{82.4}&\textbf{74.5}&\textbf{80.4}&\textbf{82.9}&\textbf{72.8}\\
				
				\midrule
		\end{tabular}}
	\end{center}
\end{table}

\subsubsection{Effect of Average Gradient Component}\label{5.4.5}

\begin{figure}[hpt]
	\centering
	\includegraphics[width=1\textwidth]{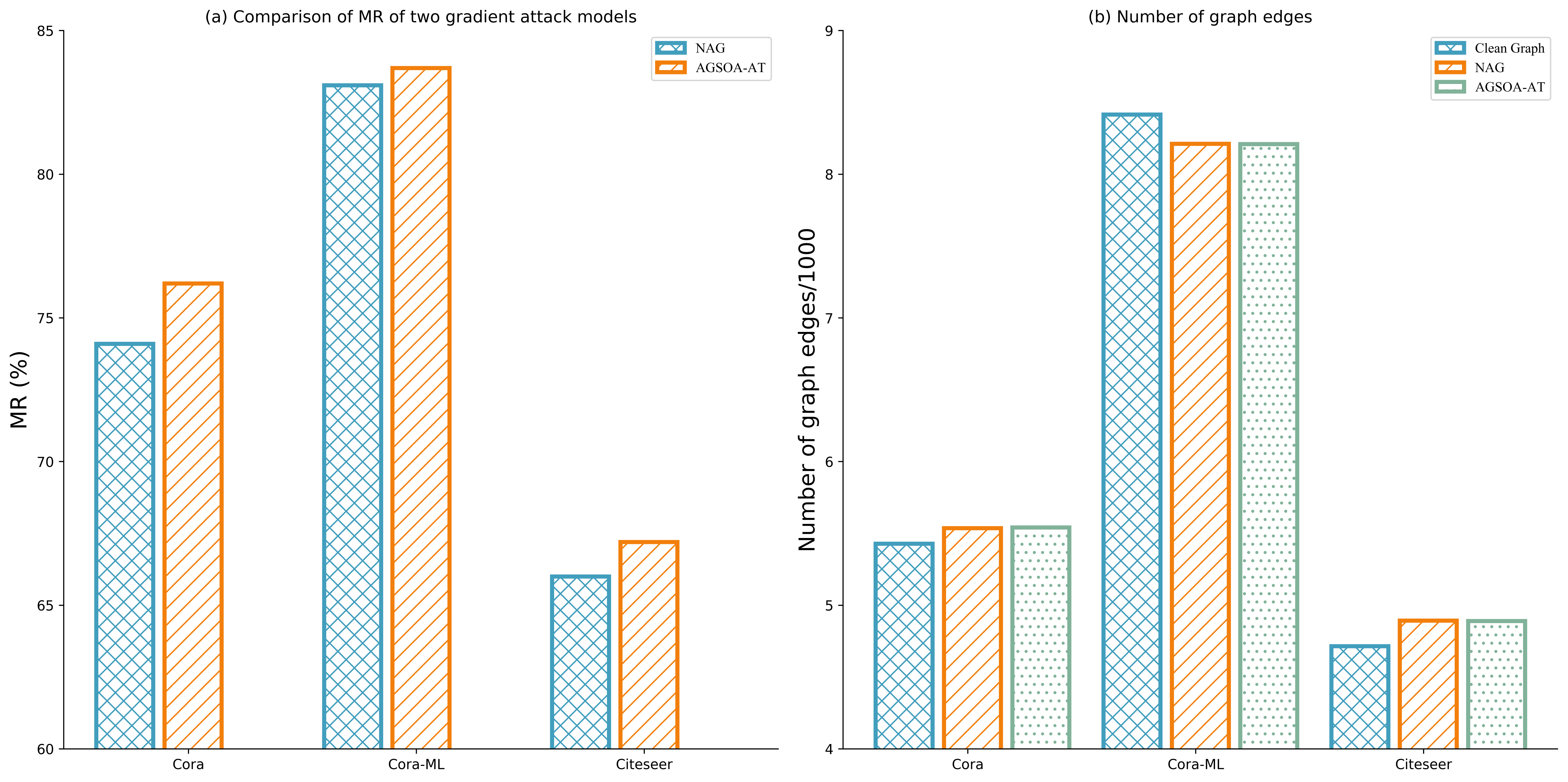}
	\caption{Effectiveness of the average gradient component. (a) is a comparison of the MR of the two gradient attack models. Where NAG is an extended model of NAG-R and is attacked using only NAG accelerated gradients. AGSOA-AT is only using the average gradient to complete the attack. (b) is a comparison of the number of edges before and after the NAG and AG attacks. To make it easier to observe the change in the number of edges, we shrink the number of edges by a factor of 1000.}\label{fig6}
\end{figure}

This section investigates the effectiveness of the average gradient component.
Fig. \ref{fig6}(a) shows the results of Average Tradient (AGSOA-AT) compared to NAG. 
AGSOA-AT is only using the average gradient to complete the attack.
In the three datasets, AGSOA-AT has a higher MR than NAG. 
Taking Cora as an example, the MR rate of AGSOA-AT is about 2$\%$ higher than that of NAG. 
The results are the same in the other datasets, which indicates that the AGSOA-AT strategy can alleviate the situation where the attack falls into a local optimum.

In particular, both NAG and AGSOA-AT have higher MR than NAG-R and AGSOA in Cora-ML.
Specifically, the MR of AGSOA-AT and AGSOA are 83.8$\%$ and 83.6$\%$, respectively.
Intuitively, NAG and AGSOA-AT operate differently for Cora-ML attacks than the other two datasets.
Fig. \ref{fig6}(b) shows the change in the number of graph edges before and after the attack. The results show  the number of perturbed graph edges is less than the clean graph after the attack in Cora-ML.
Previous research findings indicated that deleting edge operations tend to cause more damage to GNNs than adding edges in graph adversarial attacks \cite{Nettack}. Therefore, the MR of NAG and AGSOA-AT are higher than that of NAG-R and AGSOA.

Furthermore, in Fig. \ref{fig6}(b), we see that the number of modified edges is not much different between NAG and AGSOA-AT, but the AGSOA-AT  is more effective in improving the attack performance by accumulating past iteration gradients.

However, NAG and AGSOA-AT do not consider the invisibility of the attack (there is no restriction on the attack budget), which makes it difficult to implement in real attacks. Therefore, we use the graph optimization module to optimize the graph structure to improve the invisibility of the attack.

\subsubsection{Effect of Graph Optimization Component}\label{5.4.6}

The results in subsection \ref{5.4.5} demonstrate that in most cases, using only AGSOA-AT to attack GNNs is not as effective as AGSOA with the addition of a graph optimization component. The above results show that the graph optimization module is able to improve the performance of the attack. This section investigates the effect of different graph optimization methods on the performance of AGSOA.

This section uses a random optimization method (called AGSOA-R) and the graph optimization method used by NAG-R (called AGSOA-NAG) as a comparative model, where both methods first use an average gradient attack graph. In this case, AGSOA-R randomly selects the edge modification structure under the attack budget $\Delta$.

\begin{figure}[hpt]
	\centering
	\includegraphics[width=1\textwidth]{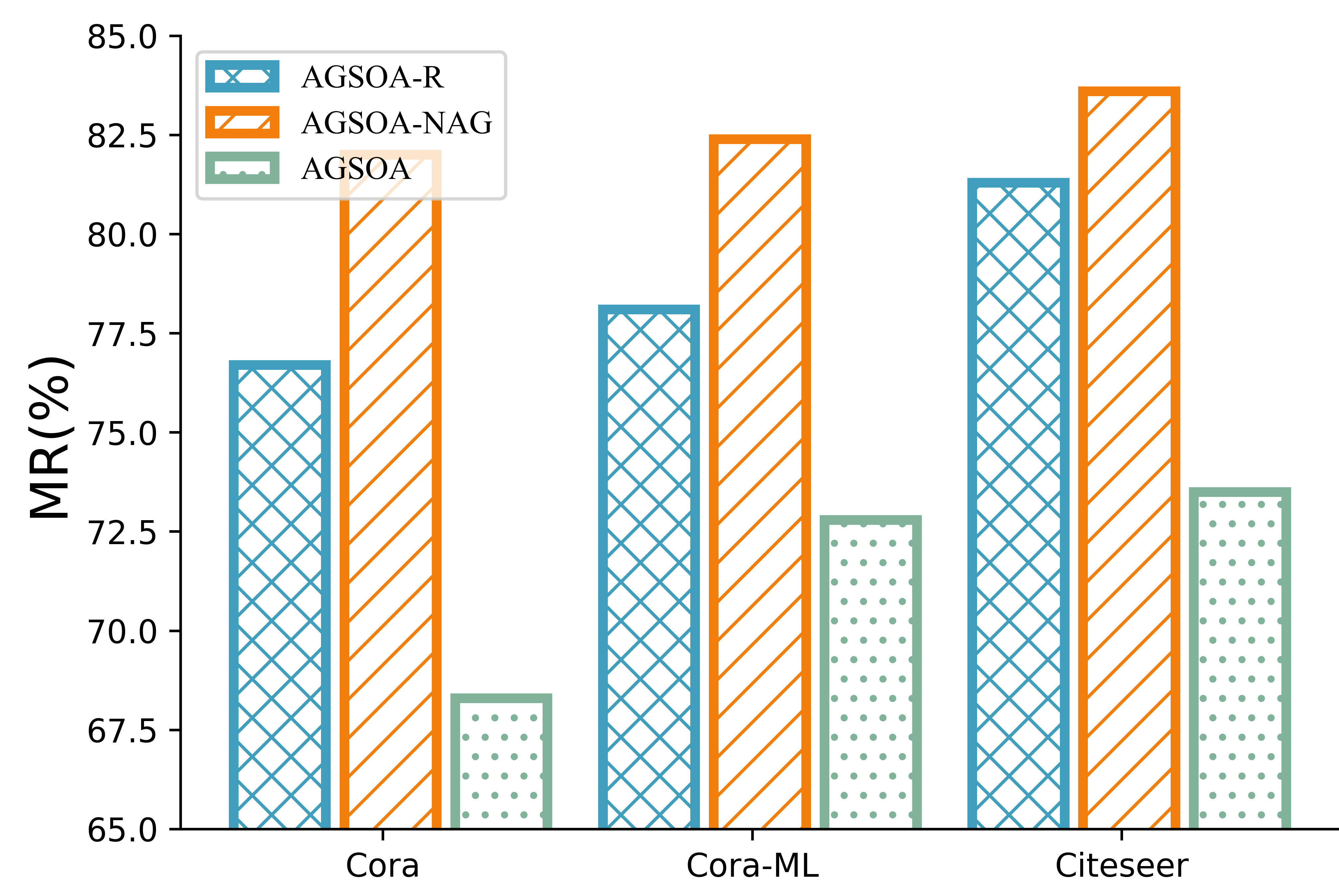}
	\caption{MR of three graph optimization attacks.}\label{fig7}
\end{figure}

Fig. \ref{fig7} The results show that our proposed graph optimization method can maximally improve the attack performance. For example, in Cora, AGSOA outperforms the two graph optimization methods AGSOA-R and AGSOA-NAG by 4.6$\%$ and 3.2$\%$, respectively. The results are the same in other datasets, which indicates that we use this optimization method to effectively destabilize and converge GNNs.

\subsubsection{Number of Target Nodes}\label{5.4.2}

\begin{figure}[hpt]
	\centering
	\includegraphics[width=1\textwidth]{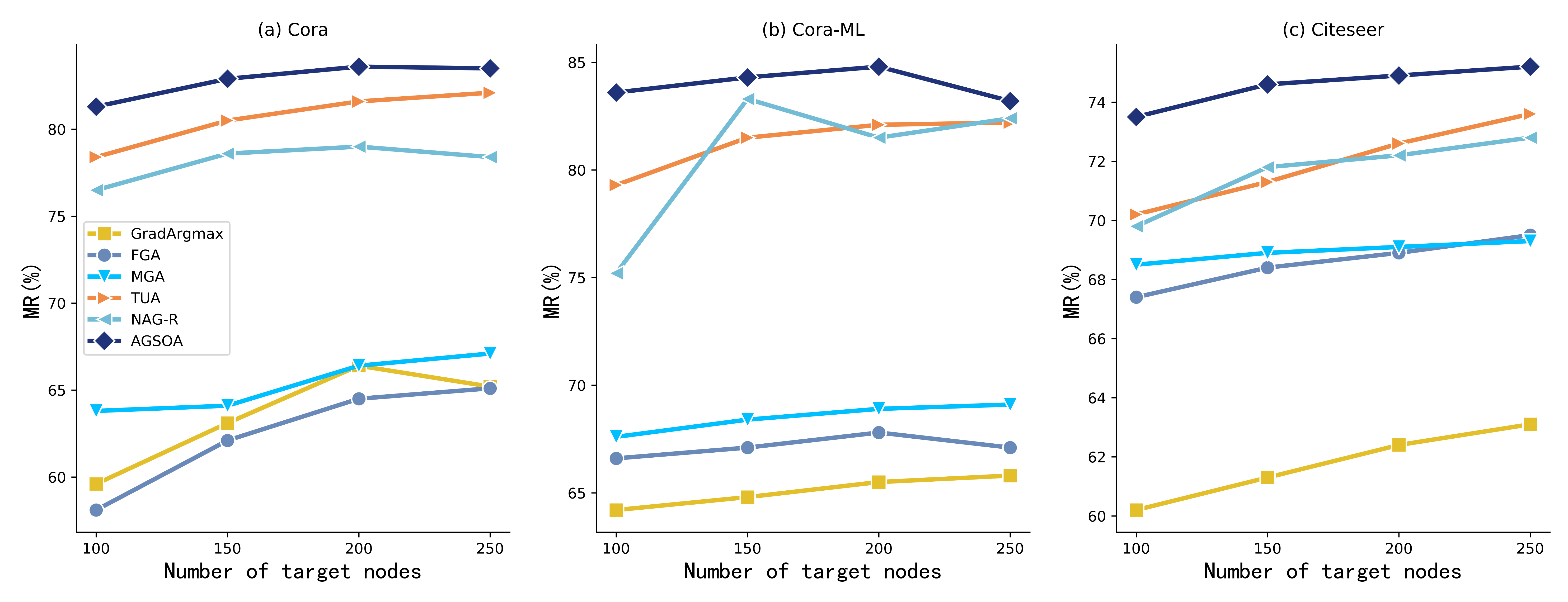}
	\caption{The attack performance under different the number of target nodes.}\label{fig3}
\end{figure}

In this section, the effect of the number of target nodes on the effectiveness of the model is investigated, and the results are shown in Fig.\ref{fig3}. Since the performance of Random is so large compared to other models, the effect of Random is not shown in Fig.\ref{fig3} in order to observe the various model variations.

Fig. \ref{fig3} reports that the attack performance increases as the number of target nodes increases. For example, in Cora, AGSOA outperforms the optimal attack model by 4.8$\%$ and 5.1$\%$ when the number of target nodes is 100 and 250, respectively. The above results show that our proposed model is able to achieve the optimal attack in different number of target nodes.

However, in Fig.\ref{fig3} b, the
FGA, NAG-R and AGSOA show a degradation of attack performance as the number of target nodes increases. 
For example, in Cora-ML, the MR of AGSOA are 84.8$\%$ and 83.2$\%$ when the number of target nodes is 200 and 250, respectively. Intuitively, Cora-ML is the densest of the three datasets, i.e., the nodes are more tightly connected to each other than the two datasets. 
When the number of attack target nodes is high, the effect of nodes influencing each other will be greater than the other datasets, therefore the performance degradation result will occur.

\subsubsection{Momentum Factor $\mu $}\label{5.4.3}

\begin{figure}[hpt]
	\centering
	\includegraphics[width=0.8\textwidth]{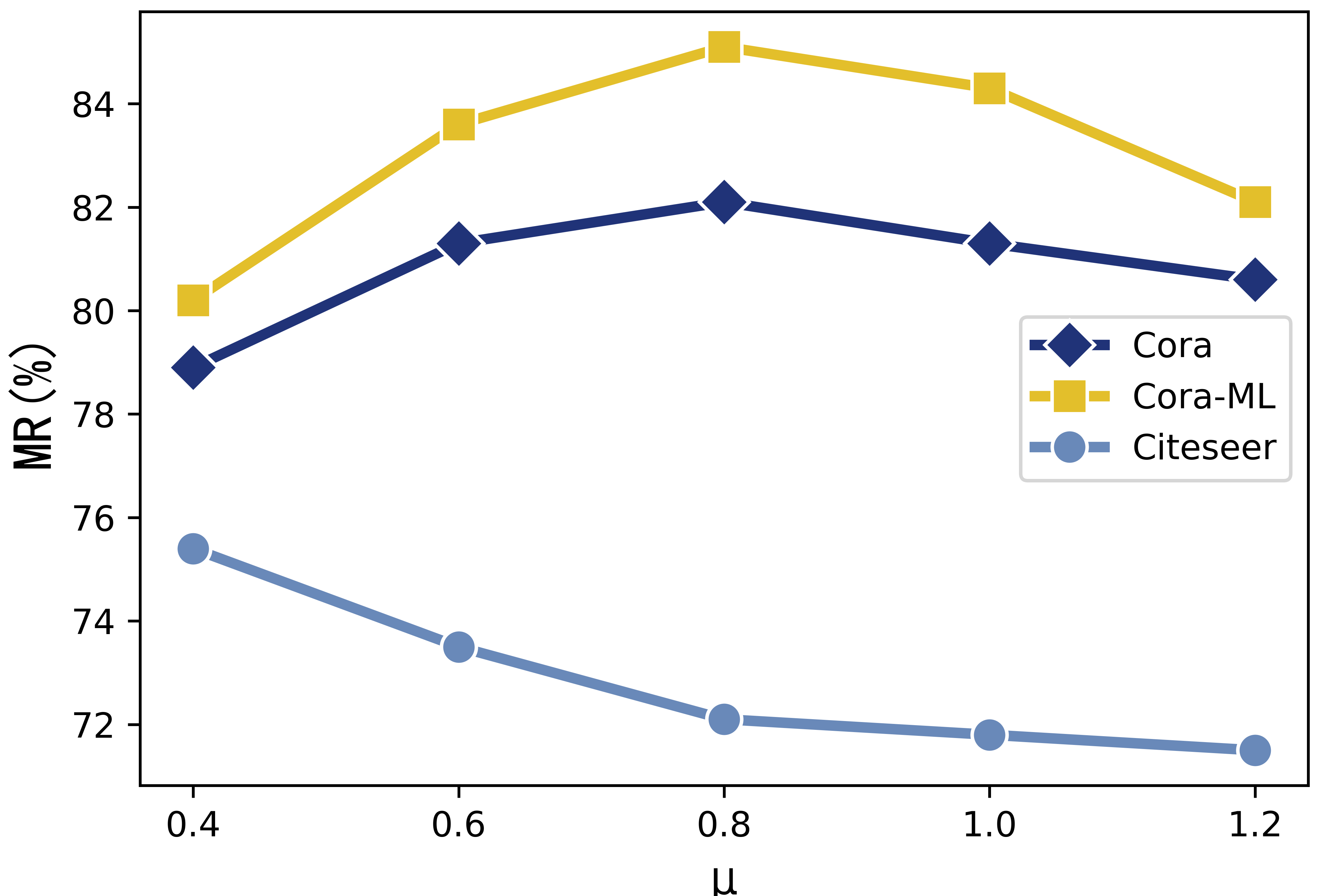}
	\caption{The attack performance under different momentum factor $\mu $.}\label{fig4}
\end{figure}

Fig.\ref{fig4} reports the effect of the momentum factor $\mu $ on the performance of the model. The larger $\mu $ is, the more the perturbations generated by the attack depend on the historical gradient.

The results in Fig.\ref{fig4} show that the history gradient can have an impact on attack performance.
In most cases, the performance of AGSOA improved as $\mu $ increased. For example, in Cora-ML, when $\mu $ are $\{$0.4, 0.6, 0.8$\}$, the misclassification rate of AGSOA are $\{$ 78.9$\%$, 81.3$\%$, 82.1$\%$ $\}$, respectively. 
In addition, Fig.\ref{fig4} shows that AGSOA reaches the optimal value when $\mu $ is different under different datasets. 
For example, in Cora, AGSOA is peaked when $\mu $ is about 0.8, whereas in Citeseer, the optimal value has been reached when $\mu $ is 0.4.

\subsubsection{Budget Parameters $\alpha $}\label{5.4.4}
\begin{figure}[hpt]
	\centering
	\includegraphics[width=0.8\textwidth]{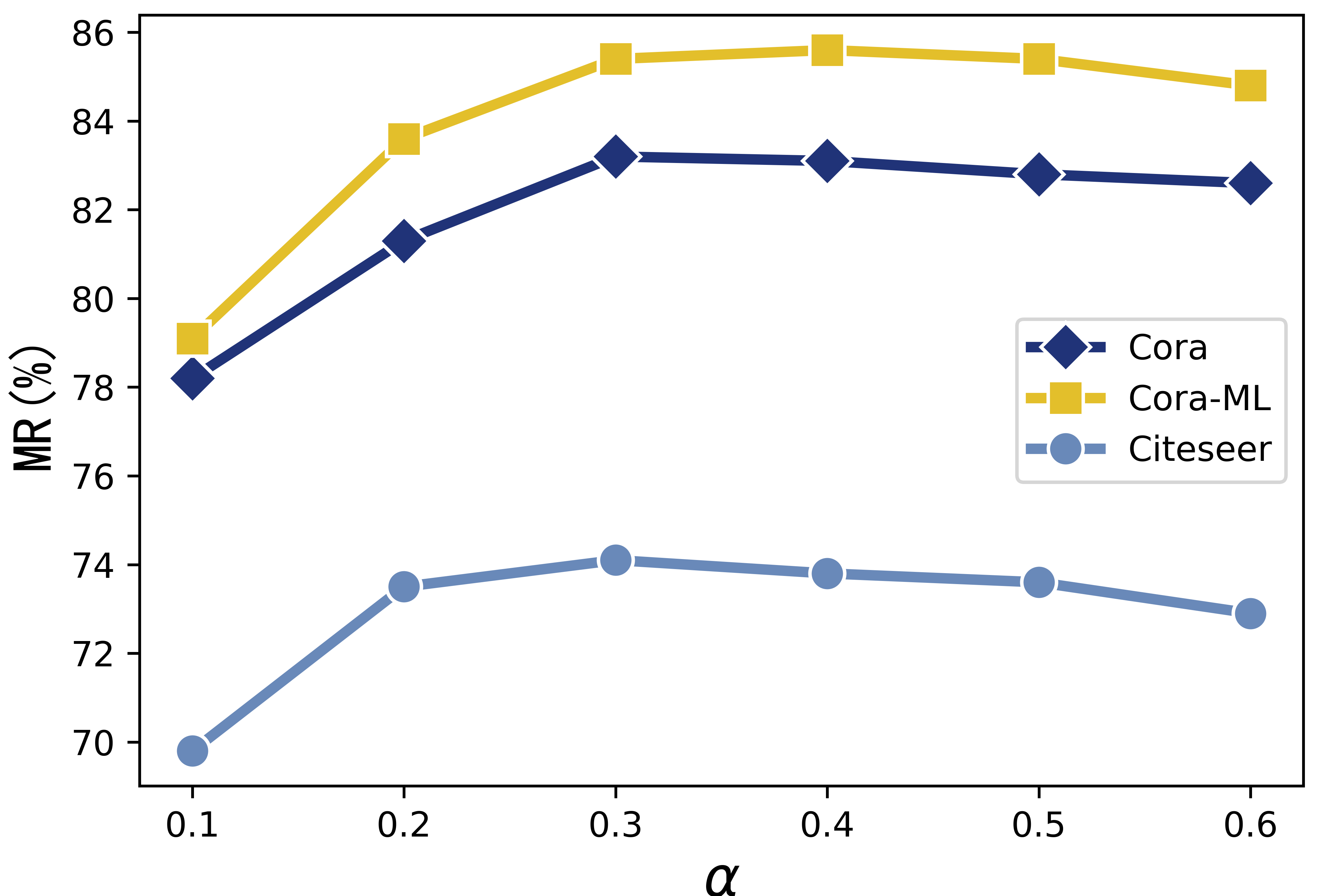}
	\caption{The model performance in different budget parameters $\alpha $.}\label{fig5}
\end{figure}

This section investigates the effect of budget parameters  $\alpha $ on the performance of AGSOA. Fig.\ref{fig5} shows that the performance of AGSOA increases and then decreases as the budget parameter  $\alpha $ increases.
For example, when $\alpha $ are $\{$0.1, 0.3, 0.6$\}$ , the MR of AGSOA are $\{$ 69.8$\%$, 74.1$\%$, 72.9$\%$ $\}$ in the Citeseer dataset. 
We think that when $\alpha $ is little, the number of AGSOA optimized structures is small, and the performance improvement of the attack is not obvious.
As $\alpha $ increases, the number of AGSOA optimized edges reaches a certain size. When all the neighbor nodes are not similar to the target nodes, the attack can successfully misclassify the target nodes. 
As $\alpha $ continues to increase, AGSOA may add or remove useless links that negatively affect the attack, so the attack performance begins to degrade.

\subsubsection{AGSOA performance under untargeted attack}\label{5.4.7}

In this section, we extend AGSOA to the untargeted attack. Unlike the targeted attack, the untargeted attack aims to operate on the entire graph structure and reduce the classification accuracy of GNNs for all nodes. The graph optimization component is designed for target nodes. Therefore, we use AGSOA-AT to verify the performance of the untargeted attack. For untargeted attacks, we strictly follow the experimental setup in NAG-NR. The attack budget $\Delta$ is set as a percentage of the perturbed edges.

From Table \ref{tab3}, we can see that our proposed model achieves the highest MR under various datasets. Using Citeseer as an example, when the attack budgets are 1$\%$ and 5$\%$, the MR of AGSOA-AT are higher than that of GradArgmax and NAG-NR by $\{$6.8$\%$, 0.5$\%$ $\}$ and $\{$9.1$\%$, 1.2$\%$ $\}$, respectively. The above results show that AGSOA is also suitable for untargeted attacks with good transferability.

\begin{table}[!ht]
	\caption{The MR($\%$) of several types of attacks under untargeted attacks. The higher the misclassification rate, the better the performance.
	}\label{tab3}%
	\begin{center}
		\resizebox{\linewidth}{!}{
			\begin{tabular}{c|ccc|ccc|ccc}
				\toprule
				Model&\multicolumn{3}{c}{Cora}&\multicolumn{3}{c}{Cora-ML}&\multicolumn{3}{c}{Citeseer}\\
				\midrule
				Budget&1$\%$&	3$\%$&5$\%$ & 1$\%$&	3$\%$&	5$\%$& 1$\%$&	3$\%$&	5$\%$\\
				\midrule
				Clean&\multicolumn{3}{c}{17.2}&\multicolumn{3}{c}{16.5}&\multicolumn{3}{c}{27.2}\\
				\midrule
				Random&17.2&17.8&18.4&16.4&16.8&17.3&27.9&28.4&	28.9\\
				GradArgmax&21.4&23.6&25.7&21.5&23.1&25.4&29.1&30.9&	31.5\\
				NAG-R&23.4&27.1&31.2&27.2&29.6&31.1&35.4&37.5&39.4\\
				AGSOA-AT&\textbf{24.4}&\textbf{27.5}&\textbf{32.4}&\textbf{28.1}&\textbf{29.1}&\textbf{31.5}&\textbf{35.9}&\textbf{38.4}&\textbf{40.6}\\
				\midrule
		\end{tabular}}
	\end{center}
\end{table}

\section{Conclusion}\label{Sec:conclusion}
In this work, we propose a targeted attack AGSOA on GNNs, which generates perturbed edges using average gradient computation and graph optimization modules.  
Specifically, AGSOA enhances the performance of the gradient attack by accumulating the average of the gradient information between the previous moment and the current moment, which avoids the attack from falling into a local optimum.
To ensure that the invisibility of the attack, we modify the edges using node similarity and node homogeneity metrics, which not only enables AGSOA to attack within a certain attack budget, but also improves the success rate of the attack.
Extensive experiments have shown that AGSOA has stronger performance and generalization capabilities than baseline methods.

\bibliographystyle{unsrt}  
\bibliography{references}

\begin{thebibliography}{10}

\bibitem{A1}
Junzhong Ji, Hao Jia, Yating Ren, and Minglong Lei.
\newblock Supervised contrastive learning with structure inference for graph
  classification.
\newblock {\em IEEE Transactions on Network Science and Engineering},
  10(3):1684--1695, MAY-JUN 2023.

\bibitem{WOS:001214105000046}
Qingju Jiao, Han Zhang, Jingwen Wu, Nan Wang, Guoying Liu, and Yongge Liu.
\newblock A simple and effective convolutional operator for node classification
  without features by graph convolutional networks.
\newblock {\em PLOS ONE}, 19(4), 2024.

\bibitem{WOS:001170500200001}
Xiaohe Li, Zide Fan, Feilong Huang, Xuming Hu, Yawen Deng, Lei Wang, and Xinyu
  Zhao.
\newblock Graph neural network with curriculum learning for imbalanced node
  classification.
\newblock {\em NEUROCOMPUTING}, 574, MAR 14 2024.

\bibitem{WOS:001175221000046}
Zeyu Wang, Jinhuan Wang, Yalu Shan, Shanqing Yu, Xiaoke Xu, Qi~Xuan, and
  Guanrong Chen.
\newblock Null model-based data augmentation for graph classification.
\newblock {\em IEEE TRANSACTIONS ON NETWORK SCIENCE AND ENGINEERING},
  11(2):1821--1833, MAR-APR 2024.

\bibitem{WOS:001124222100012}
Xiao Luo, Yusheng Zhao, Yifang Qin, Wei Ju, and Ming Zhang.
\newblock Towards semi-supervised universal graph classification.
\newblock {\em IEEE TRANSACTIONS ON KNOWLEDGE AND DATA ENGINEERING},
  36(1):416--428, JAN 2024.

\bibitem{A4}
Kai Yang, Yuan Liu, Zijuan Zhao, Xingxing Zhou, and Peijin Ding.
\newblock Graph attention network via node similarity for link prediction.
\newblock {\em The European Physical Journal B}, 96(3), MAR 2023.

\bibitem{WOS:001155057000016}
Yu~Tai, Hongwei Yang, Hui He, Xinglong Wu, and Weizhe Zhang.
\newblock A representation learning link prediction approach using line graph
  neural networks.
\newblock In Q~Liu, H~Wang, Z~Ma, W~Zheng, H~Zha, X~Chen, L~Wang, and R~Ji,
  editors, {\em PATTERN RECOGNITION AND COMPUTER VISION, PRCV 2023, PT IX},
  volume 14433 of {\em Lecture Notes in Computer Science}, pages 195--207,
  2024.

\bibitem{Transferable}
Shuiqiao Yang, Bao~Gia Doan, Paul Montague, and Olivier DeVel.
\newblock Transferable graph backdoor attack.
\newblock In {\em Proceedings of the 25th International Symposium on Research
  in Attacks, Intrusions and Defenses}, pages 321--332, 2022.

\bibitem{Nettack}
Daniel Z{\"u}gner, Amir Akbarnejad, and Stephan G{\"u}nnemann.
\newblock Adversarial attacks on neural networks for graph data.
\newblock {\em Proceedings of the 24th ACM SIGKDD international conference on
  knowledge discovery \& data mining}, pages 2847--2856, 2018.

\bibitem{liu2022towards}
Zihan Liu, Yun Luo, Lirong Wu, Zicheng Liu, and Stan~Z. Li.
\newblock Towards reasonable budget allocation in untargeted graph structure
  attacks via gradient debias.
\newblock In {\em Advances in Neural Information Processing Systems}, 2022.

\bibitem{dai2022targeted}
Jiazhu Dai, Weifeng Zhu, and Xiangfeng Luo.
\newblock A targeted universal attack on graph convolutional network by using
  fake nodes.
\newblock {\em Neural Processing Letters}, 54(4):3321--3337, 2022.

\bibitem{FGA}
Jinyin Chen, Yangyang Wu, Xuanheng Xu, Yixian Chen, Haibin Zheng, and Qi~Xuan.
\newblock Fast gradient attack on network embedding.
\newblock {\em arXiv preprint arXiv:1809.02797}, 2018.

\bibitem{wang2023revisiting}
Yongwei Wang, Yong Liu, and Zhiqi Shen.
\newblock Revisiting item promotion in gnn-based collaborative filtering: a
  masked targeted topological attack perspective.
\newblock In {\em Proceedings of the AAAI Conference on Artificial
  Intelligence}, volume~37, pages 15206--15214, 2023.

\bibitem{chen2022practical}
Yang Chen, Zhonglin Ye, Haixing Zhao, Lei Meng, Zhaoyang Wang, and Yanlin Yang.
\newblock A practical adversarial attack on graph neural networks by attacking
  single node structure.
\newblock In {\em 2022 IEEE 24th Int Conf on High Performance Computing}, pages
  143--152. IEEE, 2022.

\bibitem{lin2023exploratory}
Xixun Lin, Chuan Zhou, Jia Wu, Hong Yang, Haibo Wang, Yanan Cao, and Bin Wang.
\newblock Exploratory adversarial attacks on graph neural networks for
  semi-supervised node classification.
\newblock {\em Pattern Recognition}, 133:109042, 2023.

\bibitem{chen2022graphfool}
Jinyin Chen, Guohan Huang, Haibin Zheng, Dunjie Zhang, and Xiang Lin.
\newblock Graphfool: Targeted label adversarial attack on graph embedding.
\newblock {\em IEEE Transactions on Computational Social Systems}, 2022.

\bibitem{NAG}
Shu Zhao, Wenyu Wang, Ziwei Du, Jie Chen, and Zhen Duan.
\newblock A black-box adversarial attack method via nesterov accelerated
  gradient and rewiring towards attacking graph neural networks.
\newblock {\em IEEE Transactions on Big Data}, 2023.

\bibitem{nguyen2023poisoning}
Toan Nguyen~Thanh, Nguyen Duc~Khang Quach, Thanh~Tam Nguyen, Thanh~Trung Huynh,
  Viet~Hung Vu, Phi~Le Nguyen, Jun Jo, and Quoc Viet~Hung Nguyen.
\newblock Poisoning gnn-based recommender systems with generative
  surrogate-based attacks.
\newblock {\em ACM Transactions on Information Systems}, 41(3):1--24, 2023.

\bibitem{xing2023clean}
Xiaogang Xing, Ming Xu, Yujing Bai, and Dongdong Yang.
\newblock A clean-label graph backdoor attack method in node classification
  task.
\newblock {\em arXiv preprint arXiv:2401.00163}, 2023.

\bibitem{chen2023empirical}
Jinyin Chen, Minying Ma, Haonan Ma, Haibin Zheng, and Jian Zhang.
\newblock An empirical evaluation of the data leakage in federated graph
  learning.
\newblock {\em IEEE Transactions on Network Science and Engineering}, 2023.

\bibitem{WOS:001139144400045}
Beibei Wang, Bo~Jiang, and Chris Ding.
\newblock Fl-gnns: Robust network representation via feature learning guided
  graph neural networks.
\newblock {\em IEEE TRANSACTIONS ON NETWORK SCIENCE AND ENGINEERING},
  11(1):750--760, JAN 2024.

\bibitem{2022arXiv220812815L}
Zihan {Liu}, Ge~{Wang}, Yun {Luo}, and Stan~Z. {Li}.
\newblock {What Does the Gradient Tell When Attacking the Graph Structure}.
\newblock {\em arXiv e-prints}, August 2022.

\bibitem{AFGSM}
Jihong Wang, Minnan Luo, Fnu Suya, Jundong Li, Zijiang Yang, and Qinghua Zheng.
\newblock Scalable attack on graph data by injecting vicious nodes.
\newblock {\em Data Mining and Knowledge Discovery}, 34(5):1363--1389, SEP
  2020.

\bibitem{xu2024attacks}
Ying Xu, Michael Lanier, Anindya Sarkar, and Yevgeniy Vorobeychik.
\newblock Attacks on node attributes in graph neural networks.
\newblock {\em arXiv preprint arXiv:2402.12426}, 2024.

\bibitem{tao2024black}
Haicheng Tao, Jie Cao, Lei Chen, Hongliang Sun, Yong Shi, and Xingquan Zhu.
\newblock Black-box attacks on dynamic graphs via adversarial topology
  perturbations.
\newblock {\em Neural Networks}, 171:308--319, 2024.

\bibitem{hu2023hyperattack}
Chao Hu, Ruishi Yu, Binqi Zeng, Yu~Zhan, Ying Fu, Quan Zhang, Rongkai Liu, and
  Heyuan Shi.
\newblock Hyperattack: Multi-gradient-guided white-box adversarial structure
  attack of hypergraph neural networks, 2023.

\bibitem{MGA}
Jinyin Chen, Yixian Chen, Haibin Zheng, Shijing Shen, Shanqing Yu, Dan Zhang,
  and Qi~Xuan.
\newblock Mga: momentum gradient attack on network.
\newblock {\em IEEE Transactions on Computational Social Systems},
  8(1):99--109, 2020.

\bibitem{liu2021neighbor}
Zemin Liu, Yuan Fang, Yong Liu, and Vincent~W Zheng.
\newblock Neighbor-anchoring adversarial graph neural networks.
\newblock {\em IEEE Transactions on Knowledge and Data Engineering},
  35(1):784--795, 2021.

\bibitem{tao2021single}
Shuchang Tao, Qi~Cao, Huawei Shen, Junjie Huang, Yunfan Wu, and Xueqi Cheng.
\newblock Single node injection attack against graph neural networks.
\newblock In {\em Proceedings of the 30th ACM International Conference on
  Information \& Knowledge Management}, pages 1794--1803, 2021.

\bibitem{liu2024revisiting}
Xin Liu, Yuxiang Zhang, Meng Wu, Mingyu Yan, Kun He, Wei Yan, Shirui Pan,
  Xiaochun Ye, and Dongrui Fan.
\newblock Revisiting edge perturbation for graph neural network in graph data
  augmentation and attack.
\newblock {\em arXiv preprint arXiv:2403.07943}, 2024.

\bibitem{zhang2024maximizing}
Xiao Zhang, Peng Bao, and Shirui Pan.
\newblock Maximizing malicious influence in node injection attack.
\newblock In {\em Proceedings of the 17th ACM International Conference on Web
  Search and Data Mining}, pages 958--966, 2024.

\bibitem{chen2023feature}
Yang Chen, Zhonglin Ye, Haixing Zhao, and Ying Wang.
\newblock Feature-based graph backdoor attack in the node classification task.
\newblock {\em International Journal of Intelligent Systems}, 2023(1):5418398,
  2023.

\bibitem{chen2024imperceptible}
Yang Chen, Zhonglin Ye, Zhaoyang Wang, and Haixing Zhao.
\newblock Imperceptible graph injection attack on graph neural networks.
\newblock {\em Complex \& Intelligent Systems}, 10(1):869--883, 2024.

\bibitem{GNAI}
Junyuan Fang, Haixian Wen, Jiajing Wu, Qi~Xuan, Zibin Zheng, and Chi~K Tse.
\newblock Gani: Global attacks on graph neural networks via imperceptible node
  injections.
\newblock {\em arXiv preprint arXiv:2210.12598}, 2022.

\bibitem{sheng2021backdoor}
Yu~Sheng, Rong Chen, Guanyu Cai, and Li~Kuang.
\newblock Backdoor attack of graph neural networks based on subgraph trigger.
\newblock In {\em Collaborative Computing: Networking, Applications and
  Worksharing: 17th EAI International Conference, CollaborateCom 2021, Virtual
  Event, October 16-18, 2021, Proceedings, Part II 17}, pages 276--296.
  Springer, 2021.

\bibitem{sun2022adversarial}
Lichao Sun, Yingtong Dou, Carl Yang, Kai Zhang, Ji~Wang, S~Yu Philip, Lifang
  He, and Bo~Li.
\newblock Adversarial attack and defense on graph data: A survey.
\newblock {\em IEEE Transactions on Knowledge and Data Engineering}, 2022.

\bibitem{dai2018adversarial}
Hanjun Dai, Hui Li, Tian Tian, Xin Huang, Lin Wang, Jun Zhu, and Le~Song.
\newblock Adversarial attack on graph structured data.
\newblock In {\em International conference on machine learning}, pages
  1115--1124. PMLR, 2018.

\bibitem{yang2023gaa}
Shuxin Yang, Xiaoyang Chang, Guixiang Zhu, Jie Cao, Weiping Qin, Youquan Wang,
  and Zhendong Wang.
\newblock Gaa-ppo: A novel graph adversarial attack method by incorporating
  proximal policy optimization.
\newblock {\em Neurocomputing}, 557:126707, 2023.

\bibitem{ju2023let}
Mingxuan Ju, Yujie Fan, Chuxu Zhang, and Yanfang Ye.
\newblock Let graph be the go board: gradient-free node injection attack for
  graph neural networks via reinforcement learning.
\newblock In {\em Proceedings of the AAAI Conference on Artificial
  Intelligence}, volume~37, pages 4383--4390, 2023.

\bibitem{sharma2023task}
Kartik Sharma, Samidha Verma, Sourav Medya, Arnab Bhattacharya, and Sayan Ranu.
\newblock Task and model agnostic adversarial attack on graph neural networks.
\newblock In {\em Proceedings of the AAAI Conference on Artificial
  Intelligence}, volume~37, pages 15091--15099, 2023.

\bibitem{jain2024stronger}
Samyak Jain and Tanima Dutta.
\newblock Stronger and transferable node injection attacks.
\newblock In {\em Proceedings of the AAAI Conference on Artificial
  Intelligence}, volume~38, pages 21179--21187, 2024.

\bibitem{li2022revisiting}
Kuan Li, Yang Liu, Xiang Ao, and Qing He.
\newblock Revisiting graph adversarial attack and defense from a data
  distribution perspective.
\newblock In {\em The Eleventh International Conference on Learning
  Representations}, 2022.

\bibitem{zhang2023minimum}
Mengmei Zhang, Xiao Wang, Chuan Shi, Lingjuan Lyu, Tianchi Yang, and Junping
  Du.
\newblock Minimum topology attacks for graph neural networks.
\newblock In {\em Proceedings of the ACM Web Conference 2023}, pages 630--640,
  2023.

\bibitem{zhao2024hgattack}
He~Zhao, Zhiwei Zeng, Yongwei Wang, Deheng Ye, and Chunyan Miao.
\newblock Hgattack: Transferable heterogeneous graph adversarial attack.
\newblock {\em arXiv preprint arXiv:2401.09945}, 2024.

\bibitem{shang2023transferable}
Yu~Shang, Yudong Zhang, Jiansheng Chen, Depeng Jin, and Yong Li.
\newblock Transferable structure-based adversarial attack of heterogeneous
  graph neural network.
\newblock In {\em Proceedings of the 32nd ACM International Conference on
  Information and Knowledge Management}, pages 2188--2197, 2023.

\bibitem{GCN}
Thomas~N Kipf and Max Welling.
\newblock Semi-supervised classification with graph convolutional networks.
\newblock {\em arXiv preprint arXiv:1609.02907}, 2016.

\bibitem{liu2023enhancing}
Xuannan Liu, Yaoyao Zhong, Yuhang Zhang, Lixiong Qin, and Weihong Deng.
\newblock Enhancing generalization of universal adversarial perturbation
  through gradient aggregation.
\newblock In {\em Proceedings of the IEEE/CVF International Conference on
  Computer Vision}, pages 4435--4444, 2023.

\bibitem{WOS:001133324200016}
Chen Wan, Fangjun Huang, and Xianfeng Zhao.
\newblock Average gradient-based adversarial attack.
\newblock {\em IEEE TRANSACTIONS ON MULTIMEDIA}, 25:9572--9585, 2023.

\bibitem{mccallum2000automating}
Andrew~Kachites McCallum, Kamal Nigam, Jason Rennie, and Kristie Seymore.
\newblock Automating the construction of internet portals with machine
  learning.
\newblock {\em Information Retrieval}, 3(2):127--163, 2000.

\bibitem{bojchevski2017deep}
Aleksandar Bojchevski and Stephan G{\"u}nnemann.
\newblock Deep gaussian embedding of graphs: Unsupervised inductive learning
  via ranking.
\newblock {\em arXiv preprint arXiv:1707.03815}, 2017.

\bibitem{giles1998citeseer}
C~Lee Giles, Kurt~D Bollacker, and Steve Lawrence.
\newblock Citeseer: An automatic citation indexing system.
\newblock In {\em Proceedings of the third ACM conference on Digital
  libraries}, pages 89--98, 1998.

\bibitem{SGC}
Felix Wu, Amauri Souza, Tianyi Zhang, Christopher Fifty, Tao Yu, and Kilian
  Weinberger.
\newblock Simplifying graph convolutional networks.
\newblock In {\em International conference on machine learning}, pages
  6861--6871. PMLR, 2019.

\bibitem{ChebNet}
Will Hamilton, Zhi~Tao Ying, and Jure Leskovec.
\newblock Inductive representation learning on large graphs.
\newblock {\em Advances in neural information processing systems}, 30, 2017.

\end{thebibliography}

\end{document}